\PassOptionsToPackage{numbers, compress}{natbib}
\documentclass{article}

\usepackage[preprint]{neurips_2024}

\usepackage[utf8]{inputenc} % allow utf-8 input
\usepackage[T1]{fontenc}    % use 8-bit T1 fonts
\usepackage{hyperref}       % hyperlinks
\usepackage{url}            % simple URL typesetting
\usepackage{booktabs}       % professional-quality tables
\usepackage{amsfonts}       % blackboard math symbols
\usepackage{nicefrac}       % compact symbols for 1/2, etc.
\usepackage{microtype}      % microtypography
\usepackage{xcolor}         % colors
\usepackage{natbib}         % citation package
\usepackage{svg}
\usepackage{multirow}
\usepackage{makecell}
\usepackage{placeins}
\hypersetup{
    colorlinks=true,
    linkcolor=black,
    urlcolor=blue,
    citecolor=black
}

\newcommand{\xGenSmall}{\texttt{xGen-small}}
\newcommand{\xGenFourB}{xGen-small-4B}
\newcommand{\xGenNineB}{xGen-small-9B}

\makeatletter
\newcommand{\affiliation}[1]{%
  \gdef\@affiliation{#1}%
}

\renewcommand{\maketitle}{%
  \par
  \begingroup
    \renewcommand{\thefootnote}{\fnsymbol{footnote}}%
    \renewcommand{\@makefnmark}{\hbox to \z@{$^{\@thefnmark}$\hss}}%
    \long\def\@makefntext##1{%
      \parindent 1em\noindent
      \hbox to 1.8em{\hss $\m@th ^{\@thefnmark}$}##1%
    }%
    \thispagestyle{empty}%
    \@maketitle%         ← existing author/title block
    \vskip -1.5em
    {\centering \@affiliation \par}%
    \vskip 2em
    \@thanks%           ← your * and † footnotes
    \@notice%           ← if you have a separate notice
  \endgroup
  \let\maketitle\relax
  \let\thanks\relax
}
\makeatother

\title{xGen-small Technical Report}

\author{%
  \textbf{Erik\,Nijkamp}\thanks{These authors contributed equally.}\,\,\footnotemark[2]%
  \and \textbf{Bo\,Pang}\footnotemark[1]%
  \and \textbf{Egor\,Pakhomov}\footnotemark[1]%
  \and \textbf{Akash\,Gokul}\footnotemark[1]%
  \and \textbf{Jin\,Qu}%
  \and \textbf{Silvio\,Savarese}%
  \and \textbf{Yingbo\,Zhou}\thanks{Contact: \texttt{erik.nijkamp@salesforce.com},~\texttt{yingbo.zhou@salesforce.com},~\texttt{cxiong@salesforce.com}}%
  \and \textbf{Caiming\,Xiong}\footnotemark[2]%
}

\affiliation{%
Salesforce AI Research
\\[1em]
\includegraphics[height=1em]{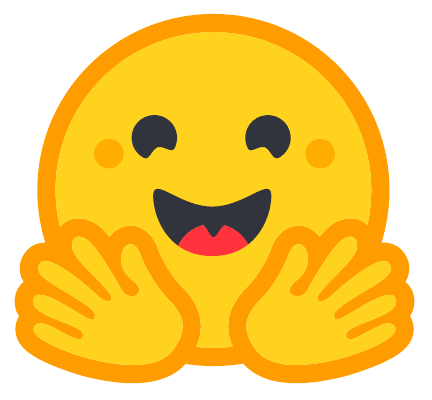}\hspace{0.3em}%
\href{https://huggingface.co/Salesforce/xgen-small}{huggingface.co/Salesforce/xgen-small}%
}

\begin{document}

\maketitle

\begin{abstract}
We introduce \xGenSmall{}, a family of 4B and 9B Transformer decoder models optimized for long‑context applications. Our vertically integrated pipeline unites domain‑balanced, frequency‑aware data curation; multi‑stage pre‑training with quality annealing and length extension to 128k tokens; and targeted post‑training via supervised fine‑tuning, preference learning, and online reinforcement learning. \xGenSmall{} delivers strong performance across various tasks, especially in math and coding domains, while excelling at long context benchmarks.
\end{abstract}

\begin{figure}[h]
    \centering
    \includegraphics[width=\linewidth]{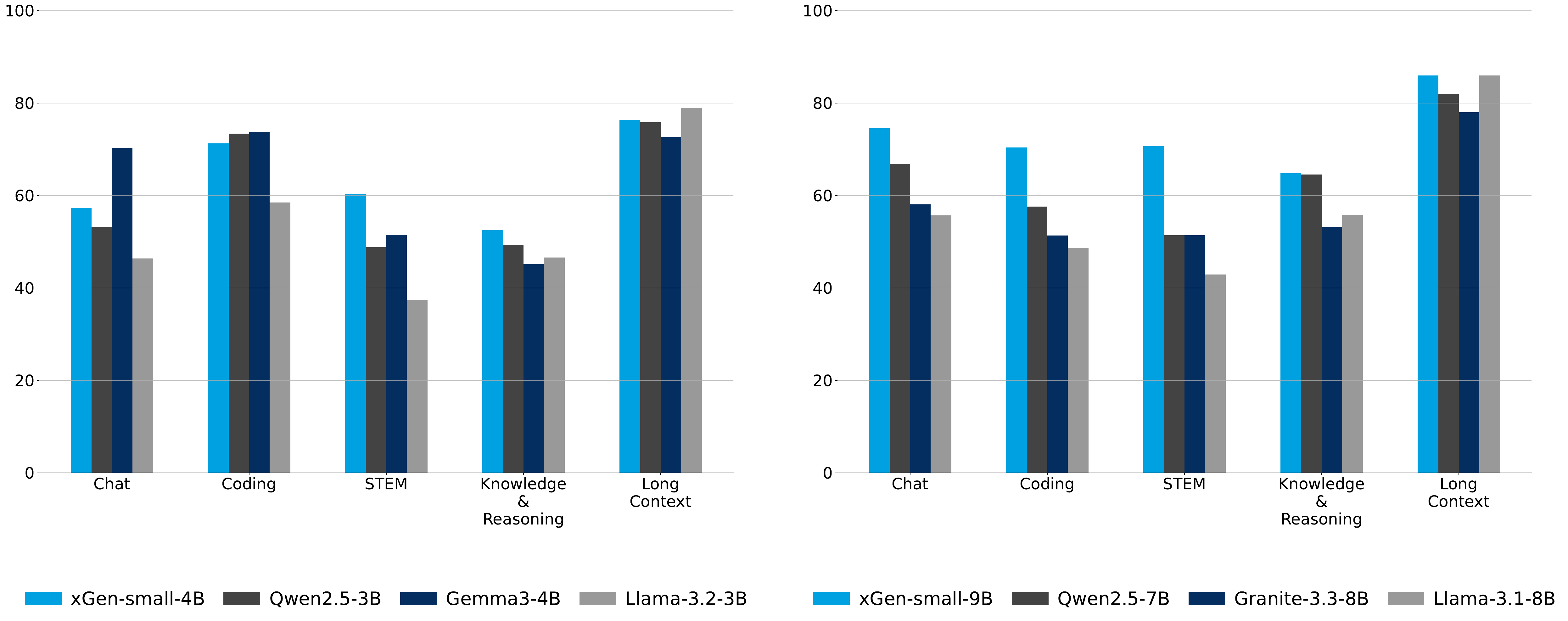}
    \caption{\textbf{Summary of xGen model performance}. We introduce the \xGenSmall{} series of language models which are available in two sizes -- \xGenFourB{} and \xGenNineB{}. These open-source models provide strong performance across a variety of tasks and scale up to 128K context length.}
    \label{fig:fig1_full_results}
\end{figure}

\section{Introduction}
\label{sec:introduction}
Deep learning–based language models have rapidly transformed natural language processing and related fields by scaling up model size and training data. However, in enterprise settings, deploying large-scale models with hundreds of billions of parameters presents significant challenges: high latency, unpredictable serving costs, elevated energy consumption, and increased risk of exposing proprietary or sensitive data. Equally important, many enterprise use cases require processing long documents, multi‐page reports, or extended conversational histories in a single pass, a capability often constrained by fixed context windows in existing LLMs.

To address these constraints, we introduce \xGenSmall{}, a family of compact Transformer decoder~\cite{vaswani2017attention} models available in two parameter scales — 4B and 9B — designed and optimized for both cost-efficient deployment and true long‐context comprehension. Instead of pursuing ever‐larger architectures, \xGenSmall{} leverages a holistic, vertically integrated pipeline that unifies:

(1) \textbf{Domain‐balanced, frequency‐aware data curation}: rigorous filtering, deduplication, and targeted up‐sampling across diverse web and domain-specific corpora to ensure both breadth and depth of high-quality content.

(2) \textbf{Pre‐training with distributional sharpening}: progressive training stages that shift from broad content coverage to high-quality subsets, sharpening the data distribution over time to improve performance across diverse tasks under aligned learning rate regimes.

(3) \textbf{Context length extension}: a two-stage methodology to extend model context from 4K to 128K tokens, utilizing tuned Rotary Position Embeddings, cross-document attention masking, and sequence parallelism to maintain training efficiency at extreme lengths.

(4) \textbf{Targeted post‐training}: supervised fine-tuning with mixed instruction datasets, off-policy preference alignment, and online reinforcement learning via Group Relative Policy Optimization (GRPO)~\cite{shao2024deepseekmath} to enhance alignment, reasoning rigor, and domain expertise.

(5) \textbf{Comprehensive evaluation}: rigorous benchmarking across general reasoning, mathematics, coding, and long‐context suites — including MMLU, GSM8K, HumanEval, and RULER — to demonstrate that \xGenSmall{} matches or outperforms similarly-sized models while retaining cost‐effective serving characteristics.

In summary, \xGenSmall{} shows that compact language models, when engineered with specialized data pipelines, efficient training strategies, dedicated length extension techniques, and targeted alignment procedures, can deliver state‐of‐the‐art performance on both standard and long‐context benchmarks without the operational overhead associated with larger models. The remainder of this report is organized as follows: Section~\ref{sec:data_curation} describes our data acquisition and curation pipeline; Section~\ref{sec:pretraining_details} details the pre‐training methodology and the context length extension approach; Section~4 outlines the post‐training alignment procedures; Section~5 covers evaluation results; and Section~6 concludes with discussion.

\section{Data-Curation}
\label{sec:data_curation}
High‑quality, domain‑balanced pre‑training data may be the single most important driver of model capability. We therefore combine large‑scale raw web data with rigorous filtering, deduplication, and targeted up-sampling to produce a data distribution that is both diverse and highly informative for enterprise use cases.

\subsection{Acquisition}
We follow the common practices for data processing pipelines: Extract documents from commonly available text corpora and split the cleaned text into self-contained documents. Each document is stored with only essential metadata (URL, crawl time, language, hash), forming a ``raw-prepared'' layer that downstream quality filtering and deduplication consume. In addition to common datasets, we incorporate relevant documents from open datasets with high concentrations of domain-specific content, such as code and math, to ensure broader and deeper domain coverage.

\subsection{Deduplication}
Deduplication is an essential step in preparing the data. We take a slightly different approach compared to prior work. Previous studies consistently show that removing large clusters of duplicate or near-duplicate documents improves model performance by reducing memorization and increasing exposure to diverse content~\cite{hernandez2022scalinglawsinterpretabilitylearning, lee2022deduplicatingtrainingdatamakes}. Yet follow‑up studies warn that pushing deduplication too far can degrade quality, probably because it strips away meaningful signals embedded in the natural frequency of documents. Although the mechanics differ — some pipelines halt deduplication part‑way (e.g., DCLM caps each document at roughly one copy per shard) ~\cite{penedo2024finewebdatasetsdecantingweb, li2025datacomplmsearchgenerationtraining}, while others keep a single canonical copy per duplicate cluster and then re‑hydrate the corpus by repeating that copy according to a coarse, bucketed log‑like recipe — often capping or even down‑weighting very large clusters~\cite{Tang2024Txt360} — the common outcome is a sub‑linear reweighting of raw frequencies, often approximating (or bounded by) a log‑scaled function rather than preserving a strict linear count.
\par
Many real‑world pipelines therefore finish deduplication in a deliberately “incomplete” state: fuzzy matching prunes the bulk of exact copies, yet enough near‑duplicates remain to implicitly up‑sample documents that appear more often in the wild. While this strategy can boost benchmark scores—higher‑frequency documents are often higher quality, as our own ablations confirm even though formal evidence in the literature is limited — it also bakes a crucial sampling decision into preprocessing. As a result, the true impact of frequency is hard to disentangle in ablation studies. We instead record natural occurrences in metadata and treat it as one quality signal among many, rather than allowing it to covertly steer dataset composition.
\par
This practice may stem from the incentive structure around releasing CommonCrawl-based datasets: new datasets are often evaluated by comparing their performance on established benchmarks, and pre-baked frequency up-sampling can produce superficially improved results. We argue that natural frequency should be explicitly tracked and used at sampling time, not silently embedded during pre-processing.
\par
Additionally, standard deduplication pipelines discard all fuzzy duplicates. If such a document is later selected for up-sampling due to high quality, we are forced to replicate a single canonical version—missing an opportunity to use alternate variants that could enrich lexical diversity. To address this, our pipeline retains the top-k fuzzy duplicates per cluster, enabling more diverse document selection during sampling.
\par
We further enrich our deduplication metadata by saving frequency signals such as the number of times a document appears across Common Crawl snapshots (indicating longevity) and across different domains (indicating breadth). Additional signals are also extracted and retained in metadata, though they are beyond the scope of this report. These are preserved alongside raw occurrence counts to inform later sampling decisions on equal footing with classifier scores and domain tags.

\subsection{Quality Filtering}
We use lightweight heuristics to discard absolutely low-quality documents not useful for training. We consider two groups of quality signals to determine which documents are retained: (1) Document natural occurrences, as described in the deduplication section, and, (2) an ensemble of FastText classifiers, each trained on a different high-quality data source.
We’ve run a large set of ablations to identify the most effective combination of these signals for selecting higher-quality documents.

\subsection{Domain Enrichment}
To strengthen reasoning capabilities, we augment the corpus with code and mathematical content: (1)
FastText domain classifiers mine additional documents from Common Crawl, and, (2)
focused web scrapers harvest content from repositories and educational sites.
We also incorporate other publicly available math/code datasets.

Additionally, as mentioned in the deduplication section, upsampling high-quality documents in the training set enhances downstream model performance. We perform upsampling separately for each quality signal, including domain-specific tags (e.g., code, math) and other quality indicators. Rather than collapsing these signals into a single composite score, we treat them independently, perform individual upsamplings, and then merge the resulting distributions. This prevents any single signal from disproportionately influencing the final data composition, thereby maintaining coverage and balance. We determine the optimal strategy through extensive ablations and careful analysis of the final dataset distribution.

\subsection{Curriculum Schedule}
Training is divided into progressive stages. Early stages use the full corpus to expose the model to breadth; later stages anneal, sharpening the distribution towards high‑quality information by focusing on the highest‑quality subset so that valuable examples are reinforced rather than overwritten. The final annealing stage comprises the smallest share of tokens but has the strictest quality threshold.
The total training budget is 8 trillion tokens.

\section{Pre-Training}
\label{sec:pretraining_details}
Pre-training of the \xGenSmall{} models involves two stages: training under distributional shifts, and length extension.

\subsection{Model}

The \xGenSmall{} models employ a standard transformer decoder architecture as introduced by~\citet{vaswani2017attention}.  Its forward pass is based on the Llama2 design~\cite{touvron2023llama}, for which we explore two parameter scales, approximately 4 billion and 9 billion parameters. To improve attention efficiency, we incorporate group query attention ~\cite{ainslie2023gqa}. The size of the vocabulary is 102,400 tokens. Positional information is encoded using Rotary Position Embeddings (RoPE) with a base frequency parameter $\theta=10,000$~\cite{su2024roformer} and a sequence length of 4,096 during pre-training. Length extension in the final stage increases $\theta=1.28\times10^{8}$ and sequence length to 131,072. 

\begin{table}[h]
  \centering
  \caption{\textbf{Key pre-training hyperparameters.}}
  \label{tab:pretrain_arch}
  \begin{tabular}{lcc}
    \toprule
    Parameter & \xGenFourB{} & \xGenNineB{}   \\
    \midrule
    Embedding size       & 4,096    & 4,096     \\
    Number of layers     & 28      & 45       \\
    Attention head size  & 128     & 128      \\
    Heads (Q/KV)         & 8/32    & 8/32     \\
    MLP hidden size      & 8,192    & 14,336    \\
    MLP activation       & silu    & silu     \\
    Sequence length      & 131,072  & 131,072 \\
    Position embedding   & RoPE    & RoPE     \\
    RoPE $\theta$        & $1.28\times10^{8}$ & $1.28\times10^{8}$ \\
    Tied embeddings      & Yes     & No       \\
    Vocabulary           & 102,400 & 102,400  \\ 
    \midrule
    Training tokens      & 8T & 8T \\
    \bottomrule
  \end{tabular}
\end{table}

\subsection{Framework}
The pre‑training of our \xGenSmall{} models is performed with our in‑house \texttt{jaxformer} library, which is written in JAX to leverage pjit/SPMD and the XLA compiler to automatically shard computation across TPU meshes across 2,048 TPU v5p cores. The library uses Haiku for the forward pass, Optax for optimization, and Orbax with TensorStore for distributed checkpointing. To increase time efficiency, fused flash attention kernels\footnote{\texttt{jax.experimental.pallas.ops.tpu.flash\_attention}} are utilized. To increase memory efficiency, we incorporate sequence parallelism for long‑context training up to 262,144 tokens.

\subsection{Training}
The pre-training recipe aims to incorporate three aspects: (1) distributional shifts from high diversity to high quality data, (2) learning rate functions for learning at high rates under unbounded tokens budgets with measurable performance gains, (3) stability in trajectory without loss spikes.

For distributional shifts, the data distribution is sharpened over the course of training from diverse data (to improve general coverage) towards high quality data (to improve performance across tasks) for programming languages (PL) and natural languages (NL). We detail the curation of this data in Section~\ref{sec:data_curation}. Specifically, we introduce four distributional shifts: (i) PL-heavy data for ease of learning under lower entropy, (ii) PL+NL mixture with high diversity to increase coverage, (iii) PL+NL mixture with gradually increasing shift towards high-quality data, (iv) PL+NL data of highest quality for final annealing inspired by~\citet{hu2024minicpm}.

For learning rate, a step-wise linear function aligned with the distributional shifts of the data distribution is employed, which follows the form of (i) warm-up, (ii) constant, (iii) slow decay, (iv) fast decay. We revised the shape of the power schedule introduced in~\citet{shen2024power} with a longer constant learning rate phase to accelerate learning. The slow decay phase allows for unbounded training under gradually improving data quality and measurable performance gains for continuous probing of the model quality. The fast decay phase allows for annealing under highest quality data as in~\citet{hu2024minicpm}.

For stable learning, the numerical precision of parameters and logits is kept in \texttt{float32}, while the remaining computation is materialized in \texttt{bfloat16}. We observed no loss spikes under our configuration.

\subsection{Length Extension}
Large language models (LLMs) are experiencing rapid advancements, enabling them to undertake increasingly complex tasks such as fixing software bugs and solving mathematical problems. These applications frequently necessitate the model's ability to process long input contexts. Consequently, the majority of current LLMs are designed with large context window capacities with 128k token context understanding as a new standard in the field. Notably, models such as Llama-3.3, Gemma-3 \cite{Kamath2025Gemma3T}, and GPT–4.1~\cite{achiam2023gpt} each purport support for context lengths up to 128k tokens. On the other hand, the Qwen2.5 \cite{Yang2024Qwen25TR} series of models, initially trained on corpora with context lengths up to 32k tokens, attain 128k context capability via the YARN \cite{peng2024yarn} approach. Similar to these open source models, the \xGenSmall{} models are endowed with the capacity to process input windows of up to 131,072 tokens.

The \xGenSmall{} models achieve this extended context capability through a two-stage context length extension training methodology. The initial checkpoint for \xGenSmall{} is limited to sequences of 4,096 tokens. In the first stage of length extension training, the maximum context is increased to 32,768 tokens, and subsequently, in the second stage, to 131,072 tokens. Inspired by the protocol outlined by~\citet{Gao2024HowTT}, and building upon its recommendations, the Rotary Position Embedding (RoPE) \cite{su2023roformerenhancedtransformerrotary} base frequency is set to $8\times10^{6}$ during the first stage of length extension, and to $1.28\times10^{8}$ in the second stage. We further incorporate cross-document attention masking, which has demonstrated enhanced performance in our ablation studies.

Although our primary objective is to extend the \xGenSmall{} models to support a 128k context window, we conduct training on even longer sequences (up to 256k tokens), thereby promoting improved performance and generalization at the 128k context length.

\section{Post-Training}
Post-training of the \xGenSmall{} models involves three sequential stages: supervised fine-tuning, off-policy preference learning, and online reinforcement learning. Each stage builds on the previous one to progressively enhance the model’s alignment, helpfulness, harmlessness, and reasoning capabilities, ultimately leading to a model that performs well across both general and high-reasoning tasks.

\subsection{Supervised Fine-Tuning}
To perform supervised fine-tuning (SFT) on \xGenSmall{} base models, we began by curating a broad, high-quality instruction dataset. The dataset is organized into three macro categories: general-purpose helpfulness, safety and harmlessness, and reasoning. The general-purpose category encompasses data from a wide range of domains, including creative writing, project management, data analysis, education, and everyday problem-solving. The safety and harmlessness category focuses on ensuring the model avoids generating harmful, biased, or unsafe content across diverse contexts. The reasoning category includes data that require advanced cognitive and analytical abilities, drawing heavily from mathematics, coding, science, and other STEM fields.

Both \xGenFourB{} and \xGenNineB{} were trained on the SFT dataset for four epochs. We used a learning rate of $5 \times 10^{-6}$ with a warm-up ratio of $10\%$. We observed consistent performance improvement over the training epochs.

Through supervised fine-tuning, the model learns a broad foundation of core capabilities. These include accurate instruction following, step-by-step reasoning, factual correctness, and desirable alignment traits such as helpfulness, honesty, and harmlessness. This stage establishes the model’s general competency and alignment with human values, preparing it for further fine-tuning.

\subsection{Off-Policy Preference Training}
Following supervised fine-tuning, we perform off-policy preference training to further align the model’s outputs with human preferences and values. In this stage, the model is trained on preference data, consisting of human-annotated rankings that indicate which responses are more preferable or aligned with user expectations. The primary goal of this phase is to improve the model’s alignment with general users by reinforcing qualities such as helpfulness, harmlessness, and informativeness. This stage focuses on shaping the model’s subjective alignment, encouraging outputs that not only follow instructions but also reflect desirable human values in open-ended situations. 

We used Direct Preference Optimization (DPO)~\cite{rafailov2023direct} to do preference alignment. The training was conducted for one epoch with $\beta=0.01$ and a learning rate of $5 \times 10^{-7}$.

Off-policy preference training improves the model’s responses across a broad spectrum of general use cases, especially in ambiguous or open-ended domains.

\subsection{Online Reinforcement Learning}
\label{sec:online_rl}
Building on preference alignment, we further train the model through online reinforcement learning to enhance its reasoning and problem-solving capabilities. In this stage, training is guided by verifiable rewards—objective signals derived from task completion accuracy or correctness, rather than subjective preference. The training data focuses on domains that demand rigorous reasoning, including mathematics, programming, and other STEM disciplines. By receiving reward signals based on whether outputs are factually or logically correct, the model incrementally improves its ability to solve complex problems, reason step-by-step, and generate verifiable, high-quality answers. 

We leverage the Group Relative Policy Optimization (GRPO)~\cite{shao2024deepseekmath} algorithm for online reinforcement learning due to its efficiency. During trajectory roll-outs, we set sampling temperature to $t=1.0$ and top-p $=1.0$. In training, learning rate is set to $10^{-6}$ and KL-Loss coefficient to $0$.

Online reinforcement learning complements preference alignment by emphasizing factual and logical rigor, leading to a model that excels in both subjective alignment and objective reasoning performance.

\section{Evaluation}
We evaluate our pre-trained and post-trained models across a comprehensive suite of tasks. These automated evaluations span various domains such as mathematics, coding, commonsense reasoning, science, and chat (for post-trained models). We also evaluate the long-context capabilities of our models using the RULER benchmark \cite{hsieh2024ruler}. To ensure a fair comparison with similarly-sized models, we evaluate other models using our pipeline. Specifically, we compare \xGenFourB{} to Llama 3.2-3B~\cite{grattafiori2024llama}, Gemma 3-4B~\cite{Kamath2025Gemma3T}, and Qwen2.5-3B~\cite{Yang2024Qwen25TR}. We compare \xGenNineB{} to Llama 3.1-8B~\cite{grattafiori2024llama}, Granite 3.3-8B~\cite{granite2024granite}, and Qwen2.5-7B~\cite{Yang2024Qwen25TR}.

% Highest scores are marked in \textbf{bold} and second-highest scores are \underline{underlined} for each evaluation task.
\begin{table*}[t!]
  \centering
  \label{tab:base_evals_9b}
  \begin{tabular}{@{}llcccc@{}}
    \toprule
    Category & Task & Llama 3.1-8B & Granite 3.3-8B & Qwen2.5-7B & \xGenNineB{} \\
    \midrule
    \multirow{7}{*}{\begin{tabular}[l]{@{}l@{}}General Knowledge \\ \& Reasoning\end{tabular}} & ARC-Challenge & 58.0 & 62.5 & \underline{63.7} & \textbf{67.4} \\
           & Big-Bench Hard & 46.3 & 46.8 & \underline{53.6} & \textbf{58.2} \\
           & HellaSwag & 81.8 & \underline{83.0} & 80.0 & \textbf{83.7} \\
           & MMLU & 65.1 & 62.7 & \textbf{74.2} & \underline{71.1} \\
           & MMLU-Pro & 32.7 & 31.3 & \textbf{43.7} & \underline{39.8} \\
           & TruthfulQA & 45.2 & \underline{52.2} & \textbf{56.4} & 48.6 \\
           & WinoGrande & 76.9 & \textbf{80.3} & 76.1 & \underline{78.6} \\
    \midrule
    \multirow{3}{*}{\begin{tabular}[l]{@{}l@{}}Math \& \\ Science\end{tabular}} & GPQA & \underline{31.9} & 30.3 & 31.4 & \textbf{32.0} \\
           & GSM8K & 55.6 & 61.4 & \underline{79.1} & \textbf{83.2} \\
           & MATH & 22.0 & 30.9 & \underline{50.2} & \textbf{52.5} \\
    \midrule
    \multirow{4}{*}{\begin{tabular}[l]{@{}l@{}}Coding\end{tabular}} & HumanEval & 37.3 & 38.9 & \textbf{55.2} & \underline{53.9} \\
           & HumanEval+ & 31.4 & 34.3 & \underline{47.7} & \textbf{47.9} \\
           & MBPP & 45.0 & 43.5 & \textbf{57.1} & \underline{50.1} \\
           & MBPP+ & 51.0 & 48.1 & \textbf{64.8} & \underline{57.6} \\
    \bottomrule
  \end{tabular}
  \caption{\textbf{Performance of pre-trained \texttt{xGen-small-9B-base}}. Highest scores are marked in \textbf{bold} and second-highest scores are \underline{underlined} for each evaluation task.}
\end{table*}
\begin{table*}[th]
  \centering
  \label{tab:base_evals_4b}
  \begin{tabular}{@{}llcccc@{}}
    \toprule
    Category & Task & Llama 3.2-3B & Gemma 3-4B & Qwen2.5-3B & \xGenFourB{} \\
    \midrule
     \multirow{7}{*}{\begin{tabular}[l]{@{}l@{}}General Knowledge \\ \& Reasoning\end{tabular}} & ARC-Challenge & 50.7 & \textbf{57.9} & 55.5 & \underline{57.3} \\
           & Big-Bench Hard & 39.1 & 40.6 & \underline{46.1} & \textbf{50.9} \\
           & HellaSwag & 76.3 & \underline{77.6} & 74.6 & \textbf{78.5} \\
           & MMLU & 56.1 & 59.5 & \textbf{65.6} & \underline{62.6} \\
           & MMLU-Pro & 25.1 & 28.1 & \textbf{32.0} & \underline{31.8} \\
           & TruthfulQA & 39.3 & 39.8 & \textbf{48.9} & \underline{42.8} \\
           & WinoGrande & 71.6 & \underline{72.3} & 70.0 & \textbf{72.7} \\
    \midrule
   \multirow{3}{*}{\begin{tabular}[l]{@{}l@{}}Math \& \\ Science\end{tabular}} & GPQA & 28.3 & \underline{29.3} & \underline{29.3} & \textbf{29.4} \\
           & GSM8K & 28.0 & 40.6 & \underline{59.1} & \textbf{71.9} \\
           & MATH & 9.0 & 25.2 & \underline{40.7} & \textbf{43.1} \\
    \midrule
    \multirow{4}{*}{\begin{tabular}[l]{@{}l@{}}Coding\end{tabular}} & HumanEval & 28.1 & 34.6 & \underline{38.3} & \textbf{42.5} \\
           & HumanEval+ & 25.6 & 28.3 & \underline{33.9} & \textbf{37.3} \\
           & MBPP & 33.2 & \underline{42.2} & 42.0 & \textbf{45.0} \\
           & MBPP+ & 38.5 & \underline{52.4} & 52.0 & \textbf{54.0} \\
    \bottomrule
  \end{tabular}
  \caption{\textbf{Performance of pre-trained \texttt{xGen-small-4B-base}}. Highest scores are marked in \textbf{bold} and second-highest scores are \underline{underlined} for each evaluation task.}
\end{table*}

\subsection{Pre-Trained Models}
Our evaluation for pre-trained models covers three main categories: general knowledge and reasoning, mathematics and science, and coding. To evaluate general knowledge and reasoning abilities, we use ARC-Challenge (25-shot) \cite{li2025datacomplmsearchgenerationtraining}, Big-Bench Hard (3-shot) \cite{suzgun-etal-2023-challenging}, HellaSwag (10-shot) \cite{zellers-etal-2019-hellaswag}, MMLU (5-shot) \cite{hendryckstest2021}, MMLU-Pro (5-shot) \cite{wang2024mmlu}, TruthfulQA (0-shot) \cite{lin-etal-2022-truthfulqa}, and WinoGrande (5-shot) \cite{10.1145/3474381}. To evaluate performance in mathematics and science, we use GPQA (0-shot) \cite{rein2024gpqa}, GSM8K (8-shot) \cite{cobbe2021training}, and MATH (4-shot) \cite{hendrycksmath2021}. To evaluate coding capabilities, we use HumanEval (0-shot) \cite{chen2021evaluating}, HumanEval+ (0-shot) \cite{liu2023your}, MBPP (0-shot) \cite{austin2021program}, and MBPP+ (0-shot) \cite{liu2023your}.

Quantitative results are presented in Tables \ref{tab:base_evals_9b} and \ref{tab:base_evals_4b}. Results show that \xGenSmall{} achieves strong performance in the 3-4B and 7-9B model classes, with the \xGenSmall{} models outperforming similarly-sized LLMs. In particular, \xGenSmall{} shows significant improvements on tasks such as Big-Bench Hard ($+4.8$ for \xGenFourB{}, $+4.6$ for \xGenNineB{}), GSM8K ($+12.8$ for \xGenFourB{}, $+4.1$ for \xGenNineB{}), and MATH ($+2.4$ for \xGenFourB{}, $+2.3$ for \xGenNineB{}). These improvements highlight the efficacy of the pre-training measures discussed in Sections \ref{sec:data_curation} and \ref{sec:pretraining_details}.

\subsection{Post-Trained Models}

% Highest scores are marked in \textbf{bold} and second-highest scores are \underline{underlined} for each evaluation task.
\begin{table*}[t]
  \centering
  \begin{tabular}{@{}llcccc@{}}
    \toprule
    Category & Task & Llama 3.1-8B & Granite 3.3-8B & Qwen2.5-7B & \xGenNineB{} \\
    \midrule
    \multirow{2}{*}{\begin{tabular}[l]{@{}l@{}}General Knowledge \\ \& Reasoning\end{tabular}} & MMLU & 68.3 & 62.7 & \textbf{72.4} & \textbf{72.4}\\
           & MMLU-Pro & 43.2 & 43.5 & \underline{56.7} & \textbf{57.3} \\
    \midrule
    \multirow{2}{*}{Chat} & Arena-Hard-v1.0 & 28.9 & 30.5 & \underline{48.1} & \textbf{60.1} \\
           & MT-Bench & 8.25 & \underline{8.57} & 8.56 & \textbf{8.90}\\
    \midrule
    \multirow{4}{*}{\begin{tabular}[l]{@{}l@{}}Math \& \\ Science\end{tabular}} & GPQA & 31.9 & \underline{35.3} & 32.6 & \textbf{45.8} \\
           & GSM8K & 84.2 & 89.4 & \underline{91.9} & \textbf{95.3} \\
           & MATH & 48.9 & 70.9 & \underline{74.6} & \textbf{91.6} \\
           & AIME 2024 & 6.7 & \underline{10.0} & 6.7 & \textbf{50.0}  \\
    \midrule
    \multirow{3}{*}{\begin{tabular}[l]{@{}l@{}}Coding\end{tabular}} & HumanEval+ & 61.6 & 65.9 & \underline{74.4} & \textbf{78.7} \\
           & MBPP+ & 55.3 & 60.3 & \textbf{68.8} & \underline{63.8} \\
           & LiveCodeBench & 10.3 & 10.3 & \underline{12.1} & \textbf{50.6} \\
    \bottomrule
  \end{tabular}
  \caption{\textbf{Performance of post-trained \texttt{xGen-small-9B-instruct}}. Highest scores are marked in \textbf{bold} and second-highest scores are \underline{underlined} for each evaluation task.}
  \label{tab:posttrained_eval_9b}
\end{table*}
\begin{table*}[t]
  \centering
  \begin{tabular}{@{}llcccc@{}}
    \toprule
    Category & Task & Llama 3.2-3B & Gemma 3-4B & Qwen2.5-3B & \xGenFourB{} \\
    \midrule
    \multirow{2}{*}{\begin{tabular}[l]{@{}l@{}}General Knowledge \\ \& Reasoning\end{tabular}} & MMLU & \underline{61.8} & 56.0 & \textbf{65.3} & 61.4 \\
           & MMLU-Pro & 31.4 & \underline{34.4} & 33.3 & \textbf{43.6} \\
    \midrule
    \multirow{2}{*}{Chat} & Arena-Hard-v1.0 & 14.7 & \textbf{57.6} & 26.9 & \underline{34.2} \\
           & MT-Bench & 7.81 & \textbf{8.29} & 7.93 & \underline{8.05} \\
    \midrule
    \multirow{4}{*}{\begin{tabular}[l]{@{}l@{}}Math \& \\ Science\end{tabular}} & GPQA & 29.7 & 31.5 & \underline{31.9} & \textbf{35.9} \\
           & GSM8K & 72.3 & \underline{90.7} & 86.3 & \textbf{92.0} \\
           & MATH & 44.6 & \underline{73.7} & 67.3 & \textbf{83.8} \\
           & AIME 2024 & 3.3 & \underline{10.0} & \underline{10.0} & \textbf{30.0} \\
    \midrule
    \multirow{3}{*}{\begin{tabular}[l]{@{}l@{}}Coding\end{tabular}} & HumanEval+ & 48.8 & 63.4 & \underline{67.7} & \textbf{74.4} \\
           & MBPP+ & 52.1 & \textbf{66.9} & \underline{61.9} & 52.6 \\
           & LiveCodeBench & 7.4 & \underline{13.6} & 10.1 & \textbf{32.6} \\
    \bottomrule
  \end{tabular}
  \caption{\textbf{Performance of post-trained \texttt{xGen-small-4B-instruct}}. Highest scores are marked in \textbf{bold} and second-highest scores are \underline{underlined} for each evaluation task.}
  \label{tab:posttrained_eval_4b}
\end{table*}

Post-trained models are evaluated to quantify performance in general knowledge and reasoning, mathematics and science, coding, and chat tasks. To evaluate general knowledge and reasoning, we use MMLU (5-shot) \cite{hendryckstest2021} and MMLU-Pro (5-shot) \cite{wang2024mmlu}. For mathematics and science evaluations, we use GPQA (0-shot) \cite{rein2024gpqa}, GSM8K (0-shot) \cite{cobbe2021training}, MATH (0-shot) \cite{hendrycksmath2021}, and AIME 2024 (0-shot). For coding evaluations, we use HumanEval+ (0-shot) \cite{liu2023your}, MBPP+ (0-shot) \cite{liu2023your} and LiveCodeBench (8/24 - 1/25) \cite{jainlivecodebench}. Finally, for chat evaluations we use Arena-Hard-v1.0 \cite{li2024crowdsourced} and MT-Bench \cite{zheng2023judging}.

Evaluation results (Tables \ref{tab:posttrained_eval_9b} and \ref{tab:posttrained_eval_4b}) highlight the strong performance of \xGenFourB{} and \xGenNineB{}. These models demonstrate competitive performance across a wide range of tasks, often leading when compared to similarly-sized models. In chat tasks, which evaluate model performance on real user queries, \xGenNineB{} significantly outperforms similarly-sized models with an Arena-Hard-v1.0 score of 60.0 ($+12.0$) and MT-Bench score of 8.9 ($+0.33$). Both \xGenNineB{} and \xGenFourB{}
also show significant improvements in math and coding domains, with notable gains on MATH ($+17$ for \xGenNineB{}, $+10.1$ for \xGenFourB{}), AIME 2024 ($+40.0$ for \xGenNineB{}, $+20.0$ for \xGenFourB{}), and LiveCodeBench ($+38.5$ for \xGenNineB{}, $+19.0$ for \xGenFourB{}). These improvements, particularly in STEM tasks, can be attributed to our post-training pipeline, specifically the online reinforcement learning step detailed in Section \ref{sec:online_rl}.

\subsection{Long Context}
\begin{figure}[t!]
    \centering
    \begin{minipage}{0.5\textwidth}
        \centering
        \includegraphics[width=\linewidth]{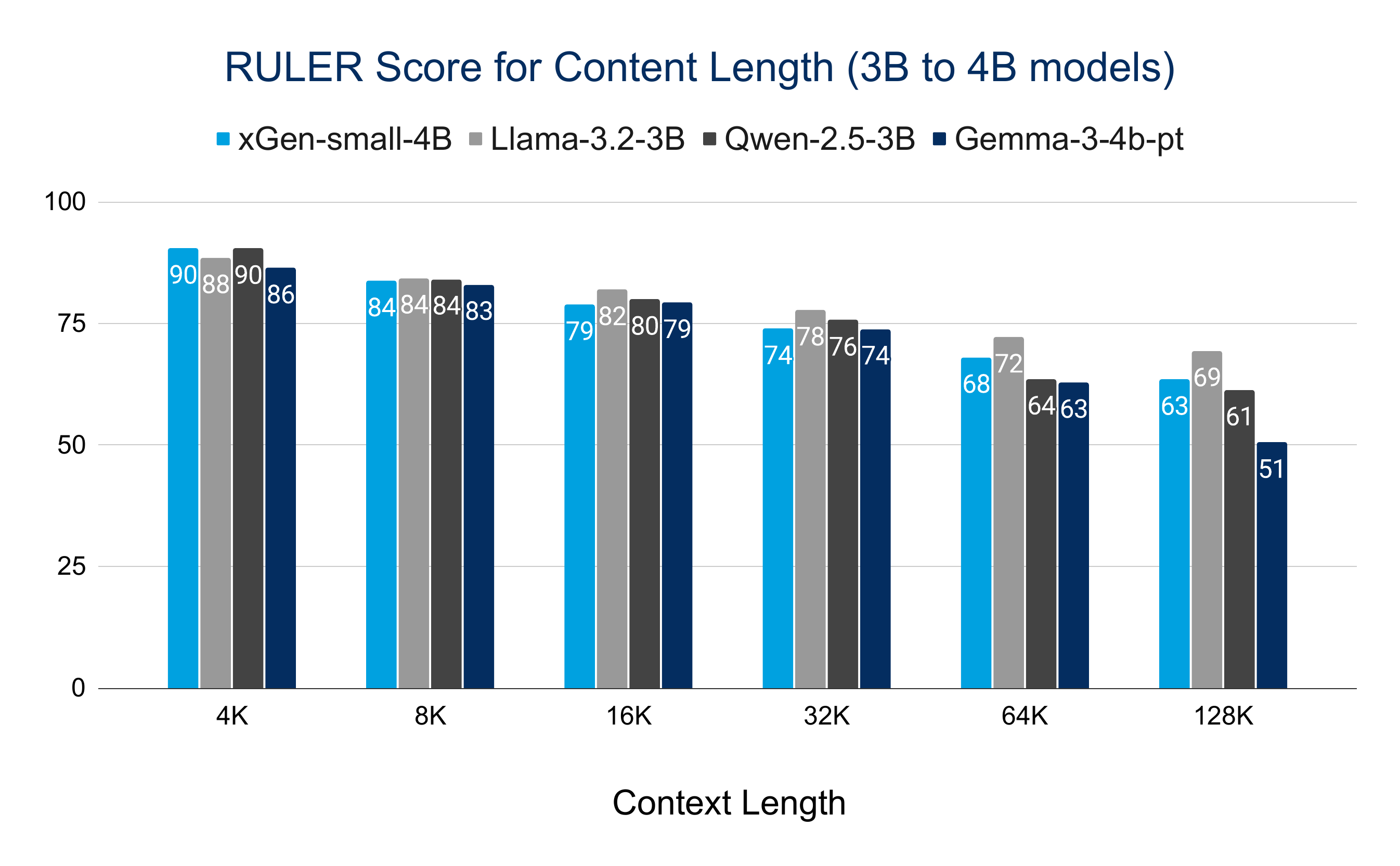}
    \end{minipage}%
    \hfill
    \begin{minipage}{0.5\textwidth}
        \centering
        \includegraphics[width=\linewidth]{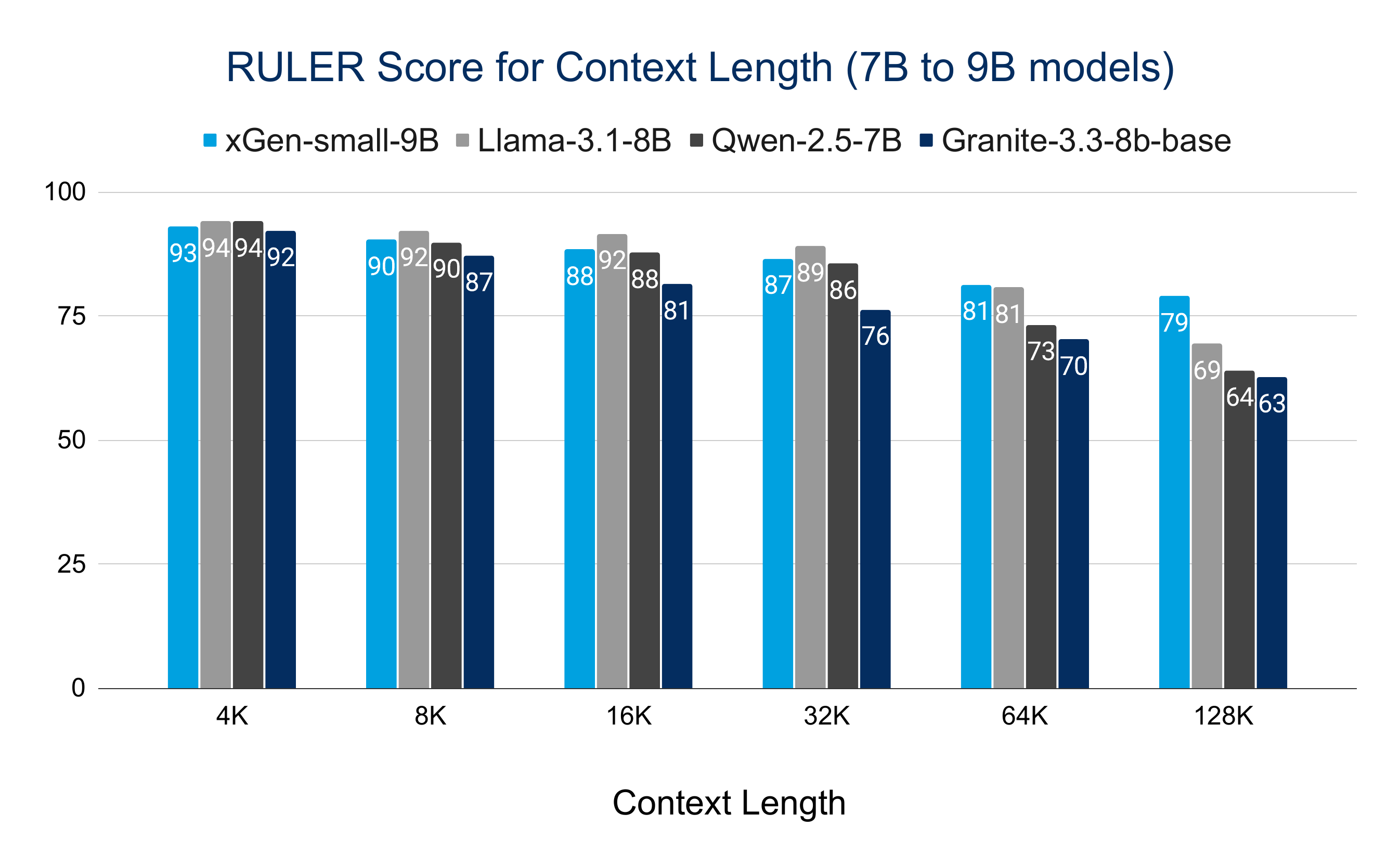}
        % e.g., (b) 7B to 9B models
    \end{minipage}
    \caption{\textbf{Detailed RULER scores for 4k-128k context length}. We compare RULER scores for models in the 3B to 4B range (left) and the 7B to 9B range (right).}
    \label{fig:ruler_scores_side_by_side}
\end{figure}

We leverage the RULER \cite{hsieh2024ruler} benchmark to measure the long context performance of \xGenSmall{} pre-trained models and their peer base models. We report the averaged RULER score in Table~\ref{tab:ruler_overall} to illustrate the overall long context capability. Figure~\ref{fig:ruler_scores_side_by_side} provides a more granular comparison across different context lengths. All models are assessed using the RULER benchmark both with and without chat templates, and we report the highest performance obtained for each model.

Empirical results indicate that the \xGenNineB{} model demonstrates superior long-context processing ability relative to other models of similar scale. As shown in Figure~\ref{fig:ruler_scores_side_by_side}, the \xGenNineB{} model maintains robust performance as the context length increases from 4k to 128k tokens. This is in stark contrast to alternative open-source models, which exhibit pronounced performance degradation at extended context lengths. For example, the RULER score of the next-best model, Llama-3.1-8B, declines by more than 10 points (on a 100-point scale) when the context length is increased from 64k to 128k tokens, whereas the \xGenNineB{} model exhibits a reduction of only approximately 2 points. The \xGenFourB{} model also achieves competitive results on the RULER test, ranking second only to Llama-3.1-8B. More detailed results are shown in Figure~\ref{fig:ruler_scores_side_by_side}. In summary, as shown in Table~\ref{tab:ruler_overall}, \xGenNineB{} and \xGenFourB{} represent the best-performing and second-best pre-trained models within their respective size class. 

\setlength{\tabcolsep}{1.5em} % for the horizontal padding
\renewcommand{\arraystretch}{1.2}
\begin{table}
\centering
    \begin{center}
    % \begin{adjustbox}{center}
        \begin{tabular}{ccc}
            \toprule
            \multicolumn{1}{c}{Size} & \multicolumn{1}{c}{Model} & \multicolumn{1}{c}{\makecell{RULER Score (4k-128k avg)}} \\
            \midrule
            \multirow{4}{*}{3-4B} & \multicolumn{1}{c}{xGen-small-4B-base} & \multicolumn{1}{c}{\underline{76.39}} \\
                                 & \multicolumn{1}{c}{Llama-3.2-3B} & \multicolumn{1}{c}{\textbf{78.98}} \\
                                 & \multicolumn{1}{c}{Qwen2.5-3B} & \multicolumn{1}{c}{75.82} \\
                                 & \multicolumn{1}{c}{Gemma-3-4B-pt} & \multicolumn{1}{c}{72.63} \\
            \midrule
            \multirow{4}{*}{7-9B} & \multicolumn{1}{c}{xGen-small-9B-base} & \multicolumn{1}{c}{\textbf{86.42}} \\
                                 & \multicolumn{1}{c}{Llama-3.1-8B} & \multicolumn{1}{c}{\underline{86.14}} \\
                                 & \multicolumn{1}{c}{Qwen2.5-7B} & \multicolumn{1}{c}{82.39} \\
                                 & \multicolumn{1}{c}{Granite-3.3-8B-base} & \multicolumn{1}{c}{78.32} \\
            \bottomrule
        \end{tabular}
    \end{center}
    \caption{\textbf{Average RULER scores across context lengths of 4k to 128k tokens}. Top score is \textbf{bolded} and the second-best score is \underline{underlined}.  We run RULER tests on all the models with and without chat template and report the best result for each model.}
    \label{tab:ruler_overall}
\end{table}

\section{Conclusion}
In this work, we introduced \xGenSmall{}, a family of compact transformer decoder models in two parameter scales (4B and 9B) engineered for true long‑context comprehension and cost‑efficient deployment. By unifying domain‑balanced, frequency‑aware data curation; multi‑stage pre‑training with progressive distributional sharpening; a two‑stage context length extension from 4K to 128K tokens; and targeted post‑training via supervised fine‑tuning, preference alignment, and online reinforcement learning (GRPO), we demonstrated that smaller models can match or exceed the performance of larger counterparts.  

Extensive evaluations on standard benchmarks (MMLU, GSM8K, HumanEval, etc.) and the RULER long‑context suite confirm that \xGenSmall{} delivers state‑of‑the‑art results in both traditional and extended‑context tasks, particularly excelling in mathematical reasoning, code generation, and processing documents up to 128K tokens. Importantly, these capabilities are achieved without the operational overhead of much larger models, making \xGenSmall{} well suited for enterprise settings where latency, cost, and privacy are paramount.

\section{Considerations}

This release is for research purposes only. Our models, datasets, and code are not specifically designed or evaluated for all downstream purposes. We strongly recommend users evaluate and address potential concerns related to accuracy, safety, and fairness before deploying this model. We encourage users to consider the common limitations of AI, comply with applicable laws, and leverage best practices when selecting use cases, particularly for high-risk scenarios where errors or misuse could significantly impact people's lives, rights, or safety. For further guidance on use cases, refer to our AUP and AI AUP.

\FloatBarrier
\bibliographystyle{unsrtnat}
\bibliography{references}

\end{document}